\documentclass[journal]{IEEEtran}
\usepackage{graphicx}
\usepackage[numbers,sort]{natbib}
\graphicspath{{images/}}
\usepackage{tikz}
\usepackage{hyperref}

\usepackage{booktabs}
\usepackage{tabularx}
\usepackage{tabulary}
\usepackage{multirow}
\newcolumntype{Y}{>{\centering\arraybackslash}X}

\usepackage{subcaption}

\usepackage{amsmath,amsfonts,bm}

\newcommand{\mathdot}{\; .}

\def\vg{{\mathbf{g}}}

\def\vm{{\mathbf{m}}}

\def\vx{{\mathbf{x}}}

\def\vz{{\mathbf{z}}}

\newcommand{\E}{\mathbb{E}}  

\def\given{{\,\vert\,}}
\newcommand{\KL}{D}
\newcommand{\KLD}[2]{\KL\big(#1 \,\|\, #2\big)}

\DeclareMathOperator*{\genmask}{mask}
\DeclareMathOperator*{\mix}{mix}
\DeclareMathOperator*{\randrect}{rand\_rect}
\DeclareMathOperator*{\filter}{filter}
\DeclareMathOperator*{\freq}{freq}
\DeclareMathOperator*{\ftop}{top}

\newcommand{\p}{p}
\newcommand{\q}{q}
\newcommand{\probsym}{\q}

\newcommand{\true}[1]{
\global\let\probsym\p
#1
\global\let\probsym\q
}

\usepackage{amssymb}

\usepackage{bm}
\usepackage{chngcntr}
\usepackage{float}
\usepackage{stfloats} 

\usepackage{xspace}
\newcommand{\fmix}{FMix\xspace}

\newcommand{\mixup}{MixUp\xspace}
\newcommand{\cutmix}{CutMix\xspace}

\newcommand{\cifar}[1]{CIFAR-{#1}\xspace}

\newcommand{\imagenet}{ImageNet\xspace}

\ifCLASSINFOpdf
\else
\fi
\begin{document}
%
\title{FMix: Enhancing Mixed Sample Data Augmentation}
%
%
%

\author{Ethan~Harris,
        Antonia~Marcu,
        Matthew~Painter,
        Mahesan~Niranjan,
        Adam~Pr{\"u}gel-Bennett,
        Jonathon~Hare%
\thanks{E. Harris, A. Marcu and M. Painter contributed equally to this work.}}
\maketitle

\begin{abstract}
Mixed Sample Data Augmentation (MSDA) has received increasing attention in recent years, with many successful variants such as \mixup and \cutmix.
By studying the mutual information between the function learned by a VAE on the original data and on the augmented data we show that \mixup distorts learned functions in a way that \cutmix does not.
We further demonstrate this by showing that \mixup acts as a form of adversarial training, increasing robustness to attacks such as Deep Fool and Uniform Noise which produce examples similar to those generated by \mixup.
We argue that this distortion prevents models from learning about sample specific features in the data, aiding generalisation performance.
In contrast, we suggest that \cutmix works more like a traditional augmentation, improving performance by preventing memorisation without distorting the data distribution.
However, we argue that an MSDA which builds on \cutmix to include masks of arbitrary shape, rather than just square, could further prevent memorisation whilst preserving the data distribution in the same way.
To this end, we propose \fmix, an MSDA that uses random binary masks obtained by applying a threshold to low frequency images sampled from Fourier space.
These random masks can take on a wide range of shapes and can be generated for use with one, two, and three dimensional data.
\fmix improves performance over \mixup and \cutmix, without an increase in training time, for a number of models across a range of data sets and problem settings, obtaining a new single model state-of-the-art result on CIFAR-10 without external data. We show that \fmix can outperform \mixup in sentiment classification tasks with one dimensional data, and provides an improvement over the baseline in three dimensional point cloud classification. Finally, we show that a consequence of the difference between interpolating MSDA such as \mixup and masking MSDA such as \fmix is that the two can be combined to improve performance even further.
Code for all experiments is provided at \url{https://github.com/ecs-vlc/FMix}.
\end{abstract}


%
\IEEEpeerreviewmaketitle

%
%
%
%
\section{Introduction}

\IEEEPARstart{R}{ecently}, a plethora of approaches to Mixed Sample Data Augmentation (MSDA) have been proposed
which obtain state-of-the-art results, particularly in classification tasks \citep{chawla2002smote,zhang2017mixup,tokozume2017learning,tokozume2018between,inoue2018data,yun2019cutmix,takahashi2019data,summers2019improved, kim2020puzzle, walawalkar2020attentive, uddin2020saliencymix}.
MSDA involves combining data samples according to some policy to create an augmented data set on which to train the model.
The policies so far proposed can be broadly categorised as either combining samples with interpolation (e.g. \mixup) or masking (e.g. \cutmix).
Traditionally, augmentation is viewed through the framework of statistical learning as Vicinal Risk Minimisation (VRM) \citep{vapnik2013nature, chapelle2001vicinal}. Given some notion of the vicinity of a data point, VRM trains with vicinal samples in addition to the data points themselves. This is the motivation for \mixup \citep{zhang2017mixup}; to provide a new notion of vicinity based on mixing data samples.
In the classical theory, validity of this technique relies on the strong assumption that the vicinal distribution precisely matches the true distribution of the data. As a result, the classical goal of augmentation is to maximally increase the data space, without changing the data distribution.
Clearly, for all but the most simple augmentation strategies, the data distribution is in some way distorted. Furthermore, there may be practical implications to correcting this, as is demonstrated in \citet{touvron2019fixing}.
In light of this, three important questions arise regarding MSDA: What is good measure of the similarity between the augmented and the original data? Why is \mixup so effective when the augmented data looks so different? If the data is distorted, what impact does this have on trained models?

To construct a good measure of similarity, we note that the data only need be `perceived' similar by the model. As such, we measure the mutual information between representations learned from the real and augmented data, thus characterising how well learning from the augmented data simulates learning from the real data. This measure shows the data-level distortion of \mixup by demonstrating that learned representations are compressed in comparison to those learned from the un-augmented data. To address the efficacy of \mixup, we look to the information bottleneck theory of deep learning \citep{tishby2015deep}. By the data processing inequality, summarised as `post-processing cannot increase information', deep networks can only discard information about the input with depth whilst preserving information about the targets. \citet{tishby2015deep} assert that more efficient generalisation is achieved when each layer maximises the information it has about the target and minimises the information it has about the previous layer. Consequently, we posit that the distortion and subsequent compression induced by \mixup promotes generalisation. Another way to view this is that compression prevents the network from learning about highly sample-specific features in the data. Regarding the impact on trained models, and again armed with the knowledge that \mixup distorts learned functions, we show that \mixup acts as a kind of adversarial training \citep{goodfellow2014explaining}, promoting robustness to additive noise. This accords with the theoretical result of \citet{perrault-archambault2020mixup} and the robustness results of \citet{zhang2017mixup}. However, we further show that MSDA does not generally improve adversarial robustness when measured as a worst case accuracy following multiple attacks as suggested by \citet{carlini2019evaluating}. Ultimately, our adversarial robustness experiments show that the distortion in the data observed by our mutual information analysis corresponds to practical differences in learned function.

In contrast to our findings regarding \mixup, our mutual information analysis shows that \cutmix causes learned models to retain a good knowledge of the real data, which we argue derives from the fact that individual features extracted by a convolutional model generally only derive from one of the mixed data points. This is further shown by our adversarial robustness results, where \cutmix is not found to promote robustness in the same way. We therefore suggest that \cutmix limits the ability of the model to over-fit by dramatically increasing the number of observable data points without distorting the data distribution,
in keeping with the original intent of VRM.
However, by restricting to only masking a square region, \cutmix imposes some unnecessary limitations. First, the number of possible masks could be much greater if more mask shapes could be used. Second, it is likely that there is still some distortion since all of the images used during training will involve a square edge.
It should be possible to construct an MSDA which uses masking similar to \cutmix whilst increasing the data space much more dramatically.
Motivated by this, we introduce \fmix, a masking MSDA that uses binary masks obtained by applying a threshold to low frequency images sampled randomly from Fourier space.
Using our mutual information measure, we show that learning with \fmix simulates learning from the real data even better than \cutmix. 
We subsequently demonstrate performance of \fmix for a range of models and tasks against a series of augmented baselines and other MSDA approaches.
\fmix obtains a new single model state-of-the-art performance on CIFAR-10 \citep{krizhevsky2009learning} without external data and improves the performance of several state-of-the-art models (ResNet, SE-ResNeXt, DenseNet, WideResNet, PyramidNet, LSTM, and Bert) on a range of problems and modalities.

In light of our experimental results, we go on to suggest that the compressing qualities of \mixup are most desirable when data is limited and learning from individual examples is easier. In contrast, masking MSDAs such as \fmix are most valuable when data is abundant.
We suggest that there is no reason to see the desirable properties of masking and interpolation as mutually exclusive. In light of these observations, we plot the performance of \mixup, \fmix, a baseline, and a hybrid policy where we alternate between batches of \mixup and \fmix, as the number of CIFAR-10 training examples is reduced. This experiment confirms our above suggestions and shows that the hybrid policy can outperform both \mixup and \fmix.

\section{Related Work}\label{sec:related}
In this section, we review the fundamentals of MSDA.
Let $p_X(x)$ denote the input data distribution. In general, we can define MSDA for a given mixing function, $\mix(X_1, X_2, \Lambda)$, where $X_1$ and $X_2$ are independent random variables on the data domain and $\Lambda$ is the mixing coefficient. Synthetic minority over-sampling \citep{chawla2002smote}, a predecessor to modern MSDA approaches, can be seen as a special case of the above where $X_1$ and $X_2$ are dependent, jointly sampled as nearest neighbours in feature space. These synthetic samples are drawn only from the minority class to be used in conjunction with the original data, addressing the problem of imbalanced data. The mixing function is linear interpolation, $\mix(x_1, x_2, \lambda) = \lambda x_1 + (1 - \lambda) x_2$, and $p_\Lambda = \mathcal{U}(0,1)$. More recently, \citet{zhang2017mixup}, \citet{tokozume2017learning}, and \citet{inoue2018data} concurrently proposed using this formulation (as \mixup, Between-Class (BC) learning, and sample pairing respectively) on the whole data set, although the choice of distribution for the mixing coefficients varies for each approach. \citet{tokozume2018between} subsequently proposed BC+, which uses a normalised variant of the mixing function. We refer to these approaches as interpolative MSDA, where, following \citet{zhang2017mixup}, we use the symmetric Beta distribution, that is $p_\Lambda = \mathrm{Beta}(\alpha, \alpha)$.

Recent variants adopt a binary masking approach \citep{yun2019cutmix,summers2019improved,takahashi2019data}. 
Let $M = \genmask(\Lambda)$ be a random variable with $\genmask(\lambda) \in \{0,1\}^{n}$ and $\mu(\genmask(\lambda))=\lambda$, that is, generated masks are binary with average value equal to the mixing coefficient.
The mask mixing function is $\mix(\vx_1, \vx_2, \vm) = \vm \odot \vx_1 + (1 - \vm) \odot \vx_2$, where $\odot$ denotes point-wise multiplication.
A notable masking MSDA which motivates our approach is \cutmix \citep{yun2019cutmix}. \cutmix is designed for two dimensional data, with 
$\genmask(\lambda)\in \{0,1\}^{w\times h}$, and uses $\genmask(\lambda) = \randrect(w\sqrt{1 - \lambda}, h\sqrt{1 - \lambda})$,
where $\randrect(r_w, r_h)\in \{0, 1\}^{w\times h}$ yields a binary mask with a shaded rectangular region of size $r_w\times r_h$ at a uniform random coordinate. 
\cutmix improves upon the performance of \mixup on a range of experiments.

In all MSDA approaches the targets are mixed in some fashion, typically to reflect the mixing of the inputs. For the typical case of classification with a cross entropy loss (and for all of the experiments in this work), the objective function is simply the interpolation between the cross entropy against each of the ground truth targets.
It could be suggested that by mixing the targets differently, one might obtain better results.
However, there are key observations from prior art which give us cause to doubt this supposition;
in particular, \citet{liang2018understanding} performed a number of experiments on the importance of the mixing ratio of the labels in \mixup. They concluded that when the targets are not mixed in the same proportion as the inputs the model can be regularised to the point of underfitting. However, despite this conclusion their results show only a mild performance change even in the extreme event that targets are mixed randomly, independent of the inputs. For these reasons, we focus only on the development of a better input mixing function for the remainder of the paper.

Attempts to explain the success of MSDAs were not only made when they were introduced, but also through subsequent empirical and theoretical studies. 
In addition to their experimentation with the targets, \citet{liang2018understanding} argue that linear interpolation of inputs limits the memorisation ability of the network.
\citet{gontijo2020affinity} propose two measures to explain the impact of augmentation on generalisation when jointly optimised: affinity and diversity.
While the former captures the shift in the data distribution as perceived by the baseline model, the latter measures the training loss when learning with augmented data. 
A more mathematical view on MSDA was adopted by \citet{guo2019mixup}, who argue that \mixup regularises the model by constraining it outside the data manifold. They point out that this could lead to reducing the space of possible hypotheses, but could also lead to generated examples contradicting original ones, degrading quality.
Upon Taylor-expanding the objective, \citet{carratino2020mixup} motivate the success of \mixup by the co-action of four different regularisation factors. 
A similar analysis is carried out in parallel by \citet{zhang2020does}.

Following \citet{zhang2017mixup}, \citet{he2019data} take a statistical learning view of MSDA, basing their study on the observation that MSDA distorts the data distribution and thus does not perform VRM in the traditional sense. They subsequently propose separating features into `minor' and `major', where a feature is referred to as `minor' if it is highly sample-specific.
Augmentations that significantly affect the distribution are said to make the model predominantly learn from `major' features.
From an information theoretic perspective, ignoring these `minor' features corresponds to increased compression of the input by the model.
Although \citet{he2019data} noted the importance of characterising the effect of data augmentation from an information perspective, they did not explore any measures that do so.
Instead, \citet{he2019data} analysed the variance in the learned representations. This is analogous to the entropy of the representation since entropy can be estimated via the pairwise distances between samples, with higher distances corresponding to both greater entropy and variance \citep{kolchinsky2017estimating}.
In proposing Manifold \mixup, \citet{pmlr-v97-verma19a} additionally suggest that \mixup works by increasing compression. The authors compute the singular values of the representations in early layers of trained networks, with smaller singular values again corresponding to lower entropy. A potential issue with these approaches is that the entropy of the representation is only an upper bound on the information that the representation has about the input.

\section{Analysis}

\begin{table}[t]
    \centering
    \caption{Mutual information between VAE latent spaces ($Z_A$) and the \cifar{10} test set ($I(Z_A;X)$), and the \cifar{10} test set as reconstructed by a baseline VAE ($I(Z_A;\hat{X})$) for VAEs trained with a range of MSDAs. \mixup prevents the model from learning about specific features in the data. Uncertainty estimates are the standard deviation following $5$ trials.}\label{tab:vaes}
    \begin{tabulary}{\linewidth}{lLLL}
    \toprule
    & $I(Z_A;X)$ & $I(Z_A;\hat{X})$ & MSE\\
    \midrule
    Baseline & 78.05$_{\pm\text{0.53}}$ &
    74.40$_{\pm\text{0.45}}$ & 
    0.256$_{\pm\text{0.002}}$\\
    \mixup{} & 70.38$_{\pm\text{0.90}}$ & 
    68.58$_{\pm\text{1.12}}$ & 0.288$_{\pm\text{0.003}}$\\
    \cutmix{} & 83.17$_{\pm\text{0.72}}$ & 79.46$_{\pm\text{0.75}}$ & 0.254$_{\pm\text{0.003}}$\\
    \bottomrule
    \end{tabulary}
\end{table}

\begin{figure}
    \centering
    \includegraphics[width=\linewidth]{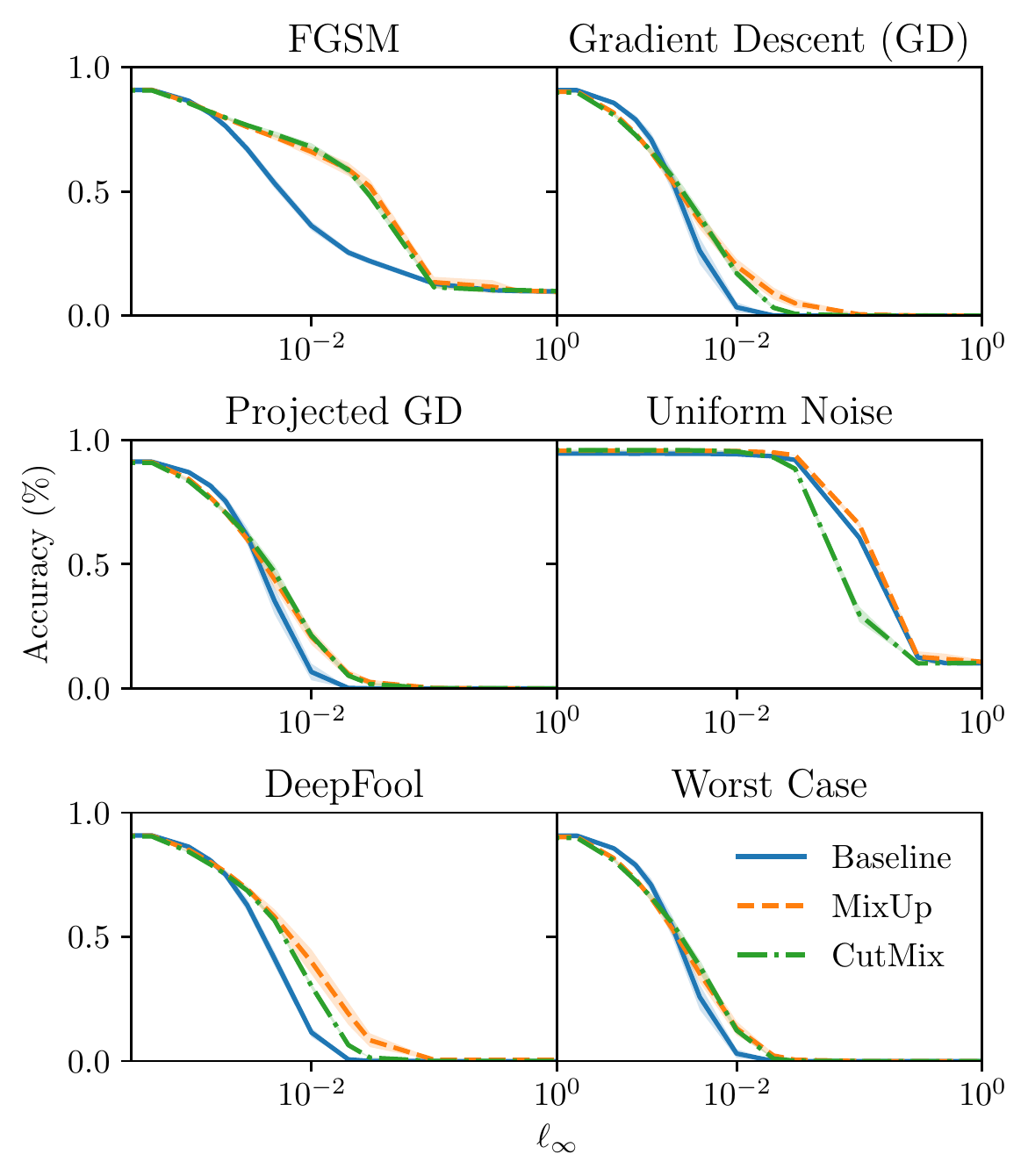}
    \caption{Robustness of PreAct-ResNet18 models trained on CIFAR-10 with standard augmentations (Baseline) and the addition of \mixup and \cutmix to the Fast Gradient Sign Method (FGSM), simple gradient descent, projected gradient descent, uniform noise, DeepFool \citep{moosavi2016deepfool}, and the worst case performance after all attacks. \mixup improves robustness to adversarial examples with similar properties to images generated with \mixup (acting as adversarial training), but MSDA does not improve robustness in general. Shaded region indicates the standard deviation following $5$ repeats.}
    \label{fig:adversarial}
\end{figure}

We now analyse both interpolative and masking MSDAs with a view to distinguishing their impact on learned representations and finding answers to the questions established in our introduction.
We first desire a measure which captures the extent to which learning about the augmented data simulates learning about the original data.
We propose training unsupervised models on real data and augmented data and then measuring the mutual information, the reduction in uncertainty about one variable given knowledge of another, between the representations they learn. To achieve this, we use Variational Auto-Encoders (VAEs)~\citep{kingma2013auto}, which provide a rich depiction of the salient or compressible information in the data~\citep{higgins2017beta}. Note that we do not expect these representations to directly relate to those of trained classifiers. Our requirement is a probabilistic model of the data, for which a VAE is well suited.

We wish to estimate the mutual information between the representation learned by a VAE from the original data set, $Z_X$, and the representation learned from some augmented data set, $Z_A$, written $I(Z_X;Z_A) = \E_{Z_X}\left[ \KLD{p_{(Z_A\given Z_X)}}{p_{Z_A}}\right]$.
This quantity acts as a good measure of the similarity between the augmented and the original data since it captures only the similarity between learnable or salient features. 
VAEs comprise an encoder, $p_{(Z\given X)}$, and a decoder, $p_{(X\given Z)}$. We impose a Normal prior on $Z$, and train the model to maximise the Evidence Lower BOund (ELBO) objective
\begin{multline}
    \mathcal{L} = \E_X \big[ \E_{Z\given X}\left[\log(p_{(X\given Z)})\right]\\
    - \KLD{p_{(Z\given X)}}{\mathcal{N}(\mathbf{0}, I)}\big]\mathdot
\end{multline}
Denoting the outputs of the decoder of the VAE trained on the augmentation as $\hat{X} = \mathit{decode}(Z_X)$, and by the data processing inequality, we have $I(Z_A;\hat{X}) \leq I(Z_A;Z_X)$ with equality when the decoder retains all of the information in $Z$.
Now, we need only observe that we already have a model of $p_{(Z_A\given X)}$, the encoder trained on the augmented data.
Estimating the marginal $p_{Z_A}$ presents a challenge as it is a Gaussian mixture. However, we can measure an alternative form of the mutual information that is equivalent up to an additive constant, and for which the divergence has a closed form solution, with
\begin{multline}
    \E_{\hat{X}}\big[ \KLD{p_{(Z_A\given \hat{X})}}{p_{Z_A}}\big] =\\
    \E_{\hat{X}}\big[\KLD{p_{(Z_A\given \hat{X})}}{\mathcal{N}(\mathbf{0}, I)}\big]\\
    - \KLD{p_{Z_A}}{\mathcal{N}(\mathbf{0}, I)}\mathdot
\end{multline}
The above holds for any choice of distribution that does not depend on $\hat{X}$. Conceptually, this states that we will always lose more information on average if we approximate $p_{(Z_A\given \hat{X})}$ with any constant distribution other than the marginal $p_{Z_A}$.
Additionally note that we implicitly minimise $\KLD{p_{Z_A}}{\mathcal{N}(\mathbf{0}, I)}$ during training of the VAE \citep{hoffman2016elbo}.
In light of this fact, we can write $I(Z_A;\hat{X})\approx \E_{\hat{X}}[\KLD{p_{(Z_A\given \hat{X})}}{\mathcal{N}(\mathbf{0}, I)}]$. We can now easily obtain a helpful upper bound of $I(Z_A;Z_X)$ such that it is bounded on both sides. Since $Z_A$ is just a function of $X$, again by the data processing inequality, we have $I(Z_A;X)\geq I(Z_A;Z_X)$. This is easy to compute since it is just the relative entropy term from the ELBO objective.

To summarise, we can compute our measure by first training two VAEs, one on the original data and one on the augmented data. We then generate reconstructions of data points in the original data with one VAE and encode them in the other. We now compute the expected value of the relative entropy between the encoded distribution and an estimate of the marginal to obtain an estimate of a lower bound of the mutual information between the representations. We then recompute this using real data points instead of reconstructions to obtain an upper bound.
Table~\ref{tab:vaes} gives these quantities for \mixup, \cutmix, and a baseline. The results show that \mixup consistently reduces the amount of information that is learned about the original data. In contrast, \cutmix manages to induce greater mutual information with the data than is obtained from just training on the un-augmented data.
Crucially, the results present concrete evidence that interpolative MSDA differs fundamentally from masking MSDA in how it impacts learned representations.

Having shown this is true for VAEs, we now wish to understand whether the finding also holds for trained classifiers. To this end, we analysed the decisions made by a classifier using Gradient-weighted Class Activation Maps (Grad-CAMs) \citep{selvaraju2017grad}. Grad-CAM finds the regions in an image that contribute the most to the network's prediction by taking the derivative of the model's output with respect to the activation maps and weighting them according to their contribution. If \mixup prevents the network from learning about highly specific features in the data we would expect more of the early features to contribute to the network output. It would be difficult to ascertain qualitatively whether this is the case. Instead, we compute the average sum of Grad-CAM heatmaps over the CIFAR-10 test set for $5$ repeats (independently trained PreAct-ResNet18 models). We obtain the following scores: baseline - $146_{\pm 5}$, \mixup{} - $162_{\pm 3}$, \cutmix{} - $131_{\pm 6}$. The result suggests that more of the early features contribute to the decisions made by \mixup trained models and that this result is consistent across independent runs.

Having established that \mixup distorts learned functions, we now seek to answer the third question from our introduction by determining the impact of data distortion on trained classifiers.
Since it is our assessment that models trained with \mixup have an altered `perception' of the data distribution, we suggest an analysis based on adversarial attacks, which involve perturbing images outside of the perceived data distribution to alter the given classification.
We perform fast gradient sign method, standard gradient descent, projected gradient descent, additive uniform noise, and DeepFool~\citep{moosavi2016deepfool} attacks over the whole CIFAR-10 test set on PreAct-ResNet18 models subject to $\ell_{\infty}$ constraints using the Foolbox library \citep{rauber2020foolboxnative,rauber2017foolbox}. The plots for the additive uniform noise and DeepFool attacks, given in Figure~\ref{fig:adversarial}, show that \mixup provides an improvement over \cutmix and the augmented baseline in this setting. 
This is because \mixup acts as a form of adversarial training \citep{goodfellow2014explaining}, equipping the models with valid classifications for images of a similar nature to those generated by the additive noise and DeepFool attacks.
In Figure~\ref{fig:adversarial}, we additionally plot the worst case robustness following all attacks as suggested by \citet{carlini2019evaluating}. These results show that the adversarial training effect of \mixup is limited and does not correspond to a general increase in robustness.
The key observation regarding these results is that there may be practical consequences to training with \mixup that are present but to a lesser degree when training with \cutmix. There may be value to creating a new MSDA that goes even further than \cutmix to minimise these practical consequences.

\section{\fmix: Improved Masking}

Our principle finding is that the masking MSDA approach works because it effectively preserves the data distribution in a way that interpolative MSDAs do not,
particularly in the perceptual space of a Convolutional Neural Network (CNN).
This derives from the fact that each convolutional neuron at a particular spatial position generally encodes information from only one of the inputs at a time. This could also be viewed as local consistency in the sense that elements that are close to each other in space typically derive from the same data point.
To the detriment of \cutmix, the number of possible examples is limited by only using square masks.
In this section we propose \fmix, a masking MSDA which maximises the number of possible masks whilst preserving local consistency.

For local consistency, we require masks that are predominantly made up of a single shape or contiguous region. We might think of this as trying to minimise the number of times the binary mask transitions from `$0$' to `$1$' or vice-versa. For our approach, we begin by sampling a low frequency grey-scale mask from Fourier space which can then be converted to binary with a threshold.
We will first detail our approach for obtaining the low frequency image before discussing our approach for choosing the threshold.
Let $Z$ denote a complex random variable with values on the domain $\mathcal{Z} = \mathbb{C}^{w\times h}$, with density
$p_{\Re(Z)} = \mathcal{N}(\mathbf{0}, \bm{I}_{w\times h})$ and $p_{\Im(Z)} = \mathcal{N}(\mathbf{0}, \bm{I}_{w\times h})$, where $\Re$ and $\Im$ return the real and imaginary parts of their input respectively. Let $\freq(w, h)\left[i, j\right]$ denote the magnitude of the sample frequency corresponding to the $i$, $j$'th bin of the $w\times h$ discrete Fourier transform.
We can apply a low pass filter to $Z$ by decaying its high frequency components. Specifically, for a given decay power $\delta$, we use
\begin{equation}
    \filter (\vz, \delta)[i, j] = \frac{\vz[i, j]}{\freq(w, h)\left[i, j\right]^\delta}\mathdot
\end{equation}
Defining $\mathcal{F}^{-1}$ as the inverse discrete Fourier transform, we can obtain a grey-scale image with
\begin{equation}
    G = \Re\big(\mathcal{F}^{-1}\big(\filter\big(Z, \delta\big)\big)\big)\mathdot
\end{equation}
All that now remains is to convert the grey-scale image to a binary mask such that the mean value is some given $\lambda$. Let $\ftop(n, \mathbf{x})$ return a set containing the top $n$ elements of the input $\mathbf{x}$. Setting the top $\lambda wh$ elements of some grey-scale image $\mathbf{g}$ to have value `$1$' and all others to have value `$0$' we obtain a binary mask with mean $\lambda$. Specifically, we have
\begin{equation}
    \genmask(\lambda, \mathbf{g})[i, j] =
    \begin{cases}
    1,& \text{if } \vg[i, j] \in \ftop(\lambda wh, \vg)\\
    0,& \text{otherwise}
    \end{cases}
    \mathdot
\end{equation}



\begin{figure}
    \centering
    \begin{tikzpicture}
        \node[] at (4.5, 0) {\includegraphics[width=0.82\linewidth,trim=2px 2px 137px 36px, clip]{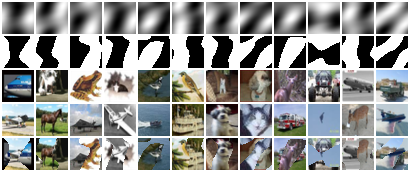}};
        \node[text width=1.2cm] at (0, 1.4) {Mask};
        \node[text width=1.2cm] at (0, 0.5) {Image 1};
        \node[text width=1.2cm] at (0, -0.4) {Image 2};
        \node[text width=1.2cm] at (0, -1.3) {\fmix};
    \end{tikzpicture}
    \caption{Example masks and mixed images from CIFAR-10 for \fmix with $\delta=3$ and $\lambda = 0.5$.}\label{fig:fmix_example}
    \label{fig:mask_visualisation}
\end{figure}
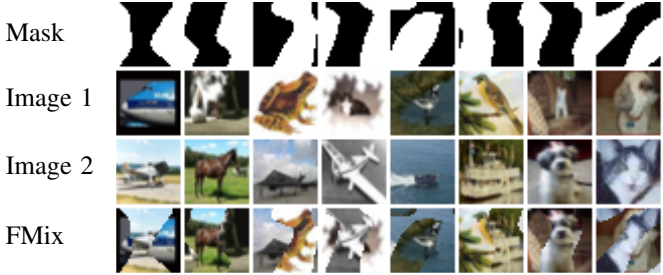

To recap, we first sample a random complex tensor for which both the real and imaginary part are independent and Gaussian. We then scale each component according to its frequency via the parameter $\delta$ such that higher values of $\delta$ correspond to increased decay of high frequency information. Next, we perform an inverse Fourier transform on the complex tensor and take the real part to obtain a grey-scale image. Finally, we set the top proportion of the image to have value `$1$' and the rest to have value `$0$' to obtain our binary mask. Although we have only considered two dimensional data here it is generally possible to create
masks with any number of dimensions.
We provide some example two dimensional masks and mixed images (with $\delta=3$ and $\lambda=0.5$) in Figure \ref{fig:fmix_example}. We can see that the space of artefacts is significantly increased, furthermore, \fmix achieves $I(Z_A;X)=83.67_{\pm\text{0.89}}$, $I(Z_A;\hat{X})=80.28_{\pm\text{0.75}}$, and $\text{MSE}=0.255_{\pm\text{0.003}}$, showing that learning from \fmix simulates learning from the un-augmented data to an even greater extent than \cutmix.







\section{Experiments}\label{sec:experiments}

We now perform a series of experiments to compare the performance of \fmix with that of \mixup, \cutmix, and augmented baselines. For each problem setting and data set, we provide exposition on the results and any relevant caveats. Throughout, we use the hyper-parameters and light augmentations (flipping, normalisation, cropping, etc.) which yield the best results in the literature for each setting.

\begin{figure}
    \centering
    \includegraphics[width=\linewidth]{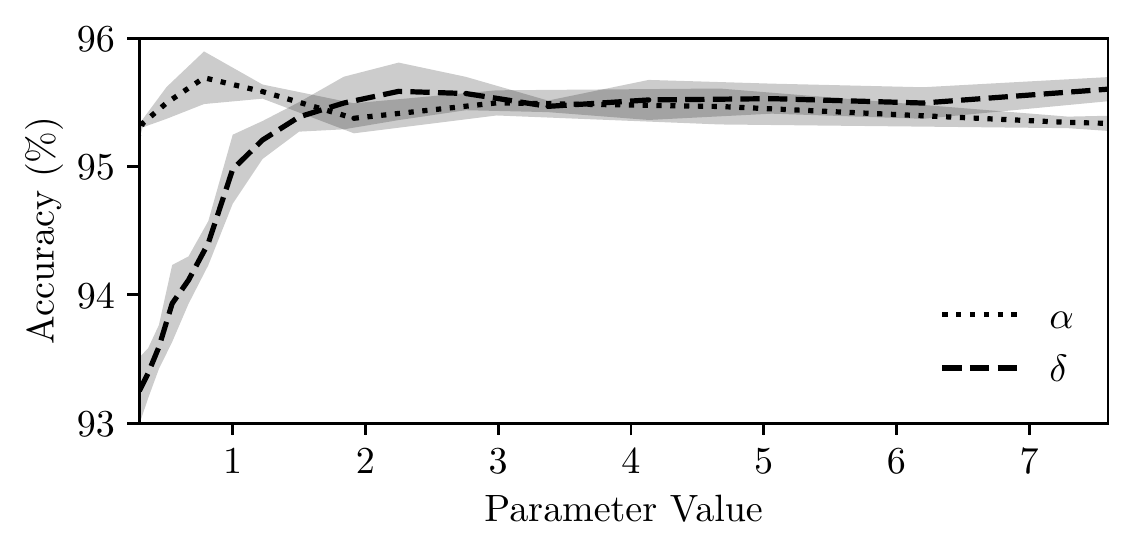}
    \caption{Ablation study showing the performance of a PreAct-ResNet18 trained on \cifar{10} with \fmix{}. Performance increases with decay power up to a point (around $\delta = 3$). Choice of $\alpha$ does not significantly impact performance.}
    \label{fig:ablation}
\end{figure}

In Figure~\ref{fig:ablation}, we perform an ablation study in order to identify sensible default values for the \fmix hyperparameters $\alpha$ and $\delta$. We see that all $\alpha$ values perform similarly, with a slight peak at $\alpha=1$, which is equivalent to sampling the mixing coefficients from a uniform distribution. Consequently, we choose this value for the majority of our experiments. For decay powers $\delta < 2$, we see decreased accuracy and $\delta \geq 2$ offering relatively consistent accuracy. We choose $\delta = 3$ for this reason and since it was found to produce large artefacts with sufficient diversity as seen Figure~\ref{fig:mask_visualisation}.   


For all experiments, we perform repeats where possible and report the average performance and standard deviation after the last epoch of training. A complete discussion of the experimental set-up can be found in Appendix~\ref{app:experimental_details} along with the standard augmentations used for all models on each data set.
In all tables, we give the best result and results that are within its margin of error in \textbf{bold}. We discuss any cases where the results obtained by us do not match the results obtained by the authors in the accompanying text, and give the authors results in parentheses. Uncertainty estimates are the standard deviation over $5$ repeats. Code for all experiments is provided at \url{https://github.com/ecs-vlc/FMix}.

\begin{table*}[t]
\centering
\caption{Image classification accuracy for our approach, \fmix, against baselines for: PreAct-ResNet18 (ResNet), WideResNet-28-10 (WRN), DenseNet-BC-190 (Dense), PyramidNet-272-200 + ShakeDrop + Fast AutoAugment (Pyramid). Parentheses indicate author quoted result.}\label{tbl:main}
\begin{tabulary}{\textwidth}{LlLLLL}
    \toprule
    Data set &  Model & Baseline & \fmix & \mixup & \cutmix\\
    \midrule
    \multirow{4}{*}{\cifar{10}}& ResNet & 94.63$_{\pm\text{0.21}}$ & \textbf{96.14}$_{\pm\text{0.10}}$ & 95.66$_{\pm\text{0.11}}$ & 96.00$_{\pm\text{0.07}}$\\
    & WRN & 95.25$_{\pm\text{0.10}}$ & 96.38$_{\pm\text{0.06}}$ & (\textbf{97.3}) \textbf{96.60}$_{\pm\text{0.09}}$ & \textbf{96.53}$_{\pm\text{0.10}}$\\
    & Dense & 96.26$_{\pm\text{0.08}}$ & \textbf{97.30}$_{\pm\text{0.05}}$ & (\textbf{97.3}) 97.05$_{\pm\text{0.05}}$ & 96.96$_{\pm\text{0.01}}$\\ 
    & Pyramid & 98.31 & \textbf{98.64} & 97.92 & 98.24\\
    \cmidrule(lr){1-6}
    \multirow{3}{*}{\cifar{100}}& ResNet & 75.22$_{\pm\text{0.20}}$ & \textbf{79.85}$_{\pm\text{0.27}}$ & (78.9) 77.44$_{\pm\text{0.50}}$ & 79.51$_{\pm\text{0.38}}$\\
    & WRN & 78.26$_{\pm\text{0.25}}$ & 82.03$_{\pm\text{0.27}}$ & (\textbf{82.5}) 81.09$_{\pm\text{0.33}}$ & 81.96$_{\pm\text{0.40}}$\\
    & Dense & 81.73$_{\pm\text{0.30}}$ & \textbf{83.95}$_{\pm\text{0.24}}$ & 83.23$_{\pm\text{0.30}}$ & 82.79$_{\pm\text{0.46}}$\\
    \cmidrule(lr){1-6}
    \multirow{3}{*}{Fashion MNIST} & ResNet & 95.70$_{\pm\text{0.09}}$ & \textbf{96.36}$_{\pm\text{0.03}}$ & 96.28$_{\pm\text{0.08}}$ & 96.03$_{\pm\text{0.10}}$\\
    & WRN & 95.29$_{\pm\text{0.17}}$ & \textbf{96.00}$_{\pm\text{0.11}}$ & 95.75$_{\pm\text{0.09}}$ & 95.64$_{\pm\text{0.20}}$ \\
    & Dense & 95.84$_{\pm\text{0.10}}$ & \textbf{96.26}$_{\pm\text{0.10}}$ & \textbf{96.30}$_{\pm\text{0.04}}$ & 96.12$_{\pm\text{0.13}}$\\
    \cmidrule(lr){1-6}
    Tiny-\imagenet & ResNet & 55.94$_{\pm\text{0.28}}$ & 61.43$_{\pm\text{0.37}}$ & 55.96$_{\pm\text{0.41}}$ & \textbf{64.08}$_{\pm\text{0.32}}$\\
    \cmidrule(lr){1-6}
    \multirow{2}{*}{Google commands} & ResNet \small($\alpha{=}\text{1.0}$) & \multirow{2}{*}{97.69$_{\pm\text{0.04}}$} & \textbf{98.59}$_{\pm\text{0.03}}$ & 98.46$_{\pm\text{0.08}}$ & 98.46$_{\pm\text{0.08}}$\\
    & ResNet \small($\alpha{=}\text{0.2}$) && \textbf{98.44}$_{\pm\text{0.06}}$ & 98.31$_{\pm\text{0.08}}$ & \textbf{98.48}$_{\pm\text{0.06}}$\\
    \cmidrule(lr){1-6}
    ModelNet10 & PointNet & 89.10$_{\pm 0.32}$ & \textbf{89.57}$_{\pm 0.44}$ & $\quad$ - & $\quad$ - \\
    \bottomrule
\end{tabulary}
\end{table*}

\begin{table*}[t]
\caption{Classification performance for a ResNet101 trained on \imagenet for 90 epochs with a batch size of 256, and evaluated on \imagenet and  \imagenet-a, adversarial examples to \imagenet. Note that \citet{zhang2017mixup} (\mixup) use a batch size of 1024 and \citet{yun2019cutmix} (\cutmix) train for 300 epochs, so these results should not be directly compared. 
}
\label{tbl:imagenet}    
\centering
\begin{tabulary}{1\linewidth}{LLLLLLLLLL}
    \toprule
      & & \multicolumn{2}{c}{Baseline} & \multicolumn{2}{c}{\fmix{}} & \multicolumn{2}{c}{\mixup{}} & \multicolumn{2}{c}{\cutmix{}}\\
      \cmidrule(lr){3-4}\cmidrule(lr){5-6}\cmidrule(lr){7-8}\cmidrule(lr){9-10}
      Data set & $\alpha$ & Top-1 & Top-5 & Top-1 & Top-5 & Top-1 & Top-5 & Top-1 & Top-5\\
      \midrule
      \multirow{2}{*}{\imagenet{}} & 1.0 & \multirow{2}{*}{77.28} & \multirow{2}{*}{93.63} & \textbf{77.42} & \textbf{93.92} & 75.89 & 93.06 & 76.92 & 93.55\\
       & 0.2 & & & \textbf{77.70} & \textbf{93.97} & 77.23 & 93.81 & 76.72 & 93.46\\
      \cmidrule(lr){1-10}
      \multirow{2}{*}{\imagenet-a} & 1.0 & \multirow{2}{*}{\hspace{0.5em}4.08} & \multirow{2}{*}{28.87} & \hspace{0.5em}7.19 & 33.65 & \hspace{0.5em}\textbf{8.69} & \textbf{34.89} & \hspace{0.5em}6.92 & 34.03\\
       & 0.2 & & & \hspace{0.5em}5.32 & 31.21 & \hspace{0.5em}5.81 & 31.43 & \hspace{0.5em}\textbf{6.08} & \textbf{31.56}\\
      \bottomrule
\end{tabulary}
\end{table*}


\subsection{Image Classification}
We first discuss image classification results on the \cifar{10/100}~\citep{krizhevsky2009learning}, Fashion MNIST~\citep{xiao2017fashion}, and Tiny-\imagenet~\citep{tinyimagenet} data sets.
Our experiments use the original version of the Fashion MNIST data set (the same data used with Random Erase \citep{zhong2017random}), which had some overlap between the train and test sets. As such, on the most recent version of the data set we would expect slightly (around $0.3\%$) lower test performance. 
We train: PreAct-ResNet18~\citep{he2016identity}, WideResNet-28-10~\citep{zagoruyko2016wide}, DenseNet-BC-190~\citep{huang2017densely} and PyramidNet-272-200~\citep{pyramidnet}. For PyramidNet, we additionally apply Fast AutoAugment~\citep{lim2019fast}, a successor to AutoAugment~\citep{cubuk2019autoaugment}, and ShakeDrop~\citep{yamada2018shakedrop} following \citet{lim2019fast}.
The results in Table~\ref{tbl:main} show that \fmix offers a significant improvement (greater than one standard deviation) over the other methods on test, with the exception of the WideResNet on \cifar{10/100} and the PreAct-ResNet on Tiny-\imagenet.
In combination with PyramidNet, \fmix achieves, to the best of our knowledge, a new state-of-the-art single model classification accuracy on \cifar{10} without use of external data. By the addition of Fast AutoAugment, this setting bares some similarity to the recently proposed AugMix~\citep{hendrycks2019augmix} which performs \mixup on heavily augmented variants of the same image.
Note that \citet{zhang2017mixup} also performed experiments with the PreAct-ResNet18, WideResNet-28-10, and DenseNet-BC-190 on \cifar{10} and \cifar{100}. There are some discrepancies between the authors results and the results obtained by our implementation. Whether any differences are significant is difficult to ascertain as no measure of deviation is provided in \citet{zhang2017mixup}.
However, since our implementation is based on the implementation from \citet{zhang2017mixup}, and most of the differences are small, we have no reason to doubt it. We speculate that these discrepancies are simply a result of random initialisation, but could also be due to differences in reporting or training configuration (we report the average terminal accuracy, some works report the best accuracy achieved at any point during training).

Next, we obtain classification results on the ImageNet Large Scale Visual Recognition Challenge (ILSVRC2012) data set~\citep{ILSVRC15}. We train a ResNet-101 on the full data set (\imagenet), additionally evaluating on \mbox{\imagenet-a}~\citep{hendrycks2019natural}, a set of natural adversarial examples to \imagenet models, to further assess adversarial robustness. We train for 90 epochs with a batch size of 256.
We perform experiments with both $\alpha=\text{1.0}$ and $\alpha=\text{0.2}$ (as this was used by \citet{zhang2017mixup}). The results, given in Table~\ref{tbl:imagenet}, show that \fmix was the only MSDA to provide an improvement over the baseline with these hyper-parameters. Note that \mixup obtains an accuracy of 78.5 in \citet{zhang2017mixup} when using a batch size of 1024. Additionally note that \mixup obtains an accuracy of 79.48 and \cutmix obtains an accuracy of 79.83 in \citet{yun2019cutmix} when training for 300 epochs. Due to hardware constraints we cannot replicate these settings and so it is not known how \fmix would compare.
On \imagenet-a, the general finding is that MSDA gives a good improvement in robustness to adversarial examples.
Interestingly, \mixup with $\alpha=\text{1.0}$ yields a lower accuracy on \imagenet but a much higher accuracy on \imagenet-a.
Since the \imagenet-a examples are chosen specifically to fool an \imagenet trained ResNet, good performance on \imagenet-a does not necessarily imply greater adversarial robustness. Instead, \imagenet-a performance can be viewed as a measure of how much the learned function differs from the baseline. These results support our argument since models which were less distorted (worse \imagenet-a performance) tended to perform better on \imagenet.

\begin{table}[t]
    \centering
    \caption{Classification performance for FMix against baselines on Bengali grapheme classification~\citep{bengali} with an SE-ResNeXt-50~\citep{xie2017aggregated,hu2018squeeze}.}
    \begin{tabularx}{\linewidth}{lXXXX}
        \toprule
        Category & Baseline & \fmix{} & \mixup{} & \cutmix{} \\
        \midrule
        Root & 92.86$_{\pm \text{0.20}}$ & \textbf{96.13}$_{\pm \text{0.14}}$ & 94.80$_{\pm \text{0.10}}$ & 95.74$_{\pm \text{0.20}}$\\
        Consonant diacritic & 96.23$_{\pm \text{0.35}}$ & \textbf{97.05}$_{\pm \text{0.23}}$ & 96.42$_{\pm \text{0.42}}$ & \textbf{96.96}$_{\pm \text{0.21}}$\\
        Vowel diacritic & 96.91$_{\pm \text{0.19}}$ & \textbf{97.77}$_{\pm \text{0.30}}$ & 96.74$_{\pm \text{0.95}}$ & 97.37$_{\pm \text{0.60}}$\\
        \cmidrule(lr){1-5}
        Grapheme & 87.60$_{\pm \text{0.45}}$ & \textbf{91.87}$_{\pm \text{0.30}}$ & 89.23$_{\pm \text{1.04}}$ & 91.08$_{\pm \text{0.49}}$\\
    \bottomrule
    \end{tabularx}
    \label{tab:bengali}
\end{table}

For a final experiment with image data, we use the Bengali.AI handwritten grapheme classification data set~\citep{bengali}, from a recent Kaggle competition.
Classifying graphemes is a multi-class problem, they consist of a root graphical form (a vowel or consonant, 168 classes) which is modified by the addition of other vowel (11 classes) or consonant (7 classes) diacritics. To correctly classify the grapheme requires classifying each of these individually, where only the root is necessarily always present. We train separate models for each sub-class, and report the individual classification accuracies and the combined accuracy (where the output is considered correct only if all three predictions are correct). We report results for 5 folds where 80\% of the data is used for training and the rest for testing. We extract the region of the image which contains the grapheme and resize to 64 $\times$ 64, performing no additional augmentation. The results for these experiments, with an SE-ResNeXt-50~\citep{xie2017aggregated,hu2018squeeze}, are given in Table~\ref{tab:bengali}. \fmix and \cutmix both clearly offer strong improvement over the baseline and \mixup, with \fmix performing significantly better than \cutmix on the root and vowel classification tasks. As a result, \fmix obtains a significant improvement when classifying the whole grapheme. In addition, note that \fmix was used in the competition by \citet{singer2020bengali} in their second place prize-winning solution. This was the best result obtained with MSDA.


\subsection{Audio Classification}
We now perform experiments on the Google Commands data set, which was created to promote deep learning research on speech recognition problems. It is comprised of 65,000 one second utterances of one of 30 words, with 10 of those words being the target classes and the rest considered unrelated or background noise.
We perform MSDA on a Mel-frequency spectrogram of each utterance.
The results for a PreAct ResNet-18 are given in Table~\ref{tbl:main}. We evaluate \fmix, \mixup, and \cutmix for the standard $\alpha=1$ used for the majority of our experiments and $\alpha=0.2$ recommended by~\citet{zhang2017mixup} for \mixup. We see in both cases that \fmix and \cutmix improve performance over \mixup outside the margin of error, with the best result achieved by \fmix with $\alpha=1$.

\subsection{Point Cloud Classification}
We now demonstrate the extension of \fmix to 3D through point cloud classification on ModelNet10~\citep{modelnet10}.
We transform the pointclouds to a voxel representation before applying a 3D \fmix mask. Table~\ref{tbl:main} reports the average median accuracy from the last 5 epochs, due to large variability in the results. Although mild, \fmix does improve performance in this setting where neither \mixup nor \cutmix can be used.

\subsection{Sentiment Analysis}
\begin{table}[t]
\centering
\caption{Classification performance of \fmix and baselines on sentiment analysis tasks. All models use $\alpha=1$ barring: Toxic Bert with $\alpha=\text{0.1}$ and IMDb CNN/BiLSTM with $\alpha=\text{0.2}$.}\label{tbl:nlp}
\begin{tabularx}{\linewidth}{lXXXX}
    \toprule
    Data set &  Model & Baseline & \fmix & \mixup \\
    \midrule
    \multirow{3}{*}{Toxic (ROC-AUC)} & CNN & 96.04$_{\pm \text{0.16}}$ & \textbf{96.80}$_{\pm\text{0.06}}$ & 96.62$_{\pm \text{0.10}}$\\
    & BiLSTM & 96.72$_{\pm\text{0.04}}$ & \textbf{97.35}$_{\pm\text{0.05}}$ & 97.15$_{\pm \text{0.06}}$\\
    & Bert & 98.22$_{\pm \text{0.03}}$ & \textbf{98.26}$_{\pm \text{0.03}}$ & $\quad$ -\\
    \cmidrule(lr){1-5}
    \multirow{2}{*}{IMDb} & CNN & 86.68$_{\pm \text{0.50}}$ & 87.31$_{\pm\text{0.34}}$ & \textbf{88.94}$_{\pm \text{0.13}}$\\
    & BiLSTM & 88.29$_{\pm \text{0.17}}$ & 88.47$_{\pm \text{0.24}}$ & \textbf{88.72}$_{\pm \text{0.17}}$\\
    \cmidrule(lr){1-5}
    \multirow{2}{*}{Yelp Binary} & CNN & 95.47$_{\pm \text{0.08}}$ & 95.80$_{\pm \text{0.14}}$ & \textbf{95.91}$_{\pm \text{0.10}}$\\
    & BiLSTM & 96.41$_{\pm \text{0.05}}$ & \textbf{96.68}$_{\pm \text{0.06}}$ & \textbf{96.71}$_{\pm \text{0.07}}$\\
    \cmidrule(lr){1-5}
    \multirow{2}{*}{Yelp Fine-grained} & CNN & 63.78$_{\pm \text{0.18}}$ & \textbf{64.46}$_{\pm \text{0.07}}$ & \textbf{64.56}$_{\pm \text{0.12}}$\\
    & BiLSTM & 62.96$_{\pm \text{0.18}}$ & \textbf{66.46}$_{\pm \text{0.13}}$ & 66.11$_{\pm \text{0.13}}$\\
    \bottomrule
\end{tabularx}
\end{table}
We can further extend the MSDA formulation for classification of one dimensional data. In Table~\ref{tbl:nlp}, we perform a series of experiments with MSDAs for the purpose of sentiment analysis. In order for MSDA to be effective, we group elements into batches of similar sequence length as is already a standard practice. This ensures that the mixing does not introduce multiple end tokens or other strange artefacts (as would be the case if batches were padded to a fixed length). The models used are: pre-trained FastText-300d~\citep{joulin2016bag} embedding followed by a simple three layer CNN~\citep{lecun1995convolutional}, the FastText embedding followed by a two layer bi-directional LSTM~\citep{hochreiter1997long}, and pre-trained Bert~\citep{devlin2018bert} provided by the HuggingFace transformers library \citep{Wolf2019HuggingFacesTS}. For the LSTM and CNN models we compare \mixup and \fmix with a baseline. For the Bert fine-tuning we do not compare to \mixup as the model input is a series of tokens, interpolations between which are meaningless. We first report results on the Toxic Comments~\citep{toxic} data set, a Kaggle competition to classify text into one of 6 classes. For this data set we report the ROC-AUC metric, as this was used in the competition. Note that these results are computed over the whole test set and are therefore not comparable to the competition scores, which were computed over a subset of the test data. In this setting, both \mixup and \fmix provide an improvement over the baseline, with \fmix consistently providing a further improvement over \mixup. The improvement when fine-tuning Bert with \fmix is outside the margin of error of the baseline, but mild in comparison to the improvement obtained in the other settings. We additionally report results on the IMDb~\citep{maas2011learning}, Yelp binary, and Yelp fine-grained~\citep{zhang2015character} data sets. For the IMDb data set, which has one tenth of the number of examples, we found $\alpha=0.2$ to give the best results for both MSDAs. Here, \mixup provides a clear improvement over both \fmix and the baseline for both models. This suggests that \mixup may perform better when there are fewer examples.

\subsection{Combining MSDAs}

\begin{figure}
    \centering
    \includegraphics[width=\linewidth]{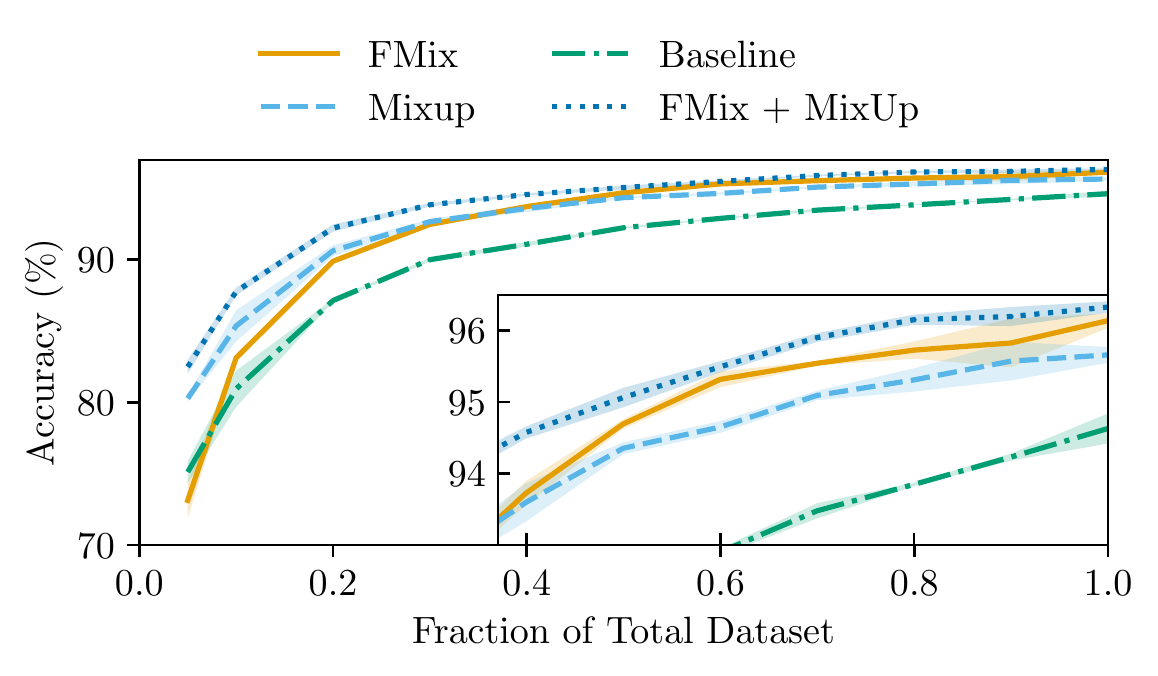}
    \captionof{figure}{\cifar{10} performance for a PreAct-ResNet18 as we change the amount of training data.}
    \label{fig:split_ablate}
\end{figure}

We have established through our analysis that models trained with interpolative MSDA perform a different function to models trained with masking. We now wish to understand whether the benefits of interpolation and masking are mutually exclusive. We therefore perform experiments with simultaneous action of multiple MSDAs, alternating their application per batch with a PreAct-ResNet18 on \cifar{10}.
A combination of interpolation and masking, particularly \fmix{}+\mixup ($96.30_{\pm0.08}$), gives the best results, with \cutmix{}+\mixup performing slightly worse ($96.26_{\pm0.04}$).
In contrast, combining \fmix and \cutmix gives worse results ($95.85_{\pm0.1}$) than using either method on its own.
For a final experiment, we note that our results suggest that interpolation performs better when there is less data available (e.g. the IMDb data set) and that masking performs better when there is more data available (e.g. \imagenet and the Bengali.AI data set). This finding is supported by our analysis since it is always easier for the model to learn specific features, and so we would naturally expect that preventing this is of greater utility, when there is less data.
We confirm this empirically by varying the size of the \cifar{10} training set and training with different MSDAs in Figure~\ref{fig:split_ablate}. Notably, the \fmix{}+\mixup policy obtains superior performance irrespective of the amount of available data.

\section{Conclusions and Future Work}

In this paper we have introduced \fmix, a masking MSDA that improves classification performance for a series of models, modalities, and dimensionalities.
We believe the strength of masking methods resides in preserving local features and we improve upon existing approaches by increasing the number of possible mask shapes.
We have verified this intuition through a novel analysis.
Our analysis shows that interpolation causes models to encode more general features, whereas masking causes models to encode the same information as when trained with the original data whilst eliminating memorisation.
Our preliminary experiments suggest that combining interpolative and masking MSDA could improve performance further, although further work is needed to fully understand this phenomenon.
Future work should also look to expand on the finding that masking MSDA works well in combination with Fast AutoAugment~\citep{lim2019fast}, perhaps by experimenting with similar methods like AutoAugment~\citep{cubuk2019autoaugment} or RandAugment~\citep{cubuk2019randaugment}.

\appendices
\section{Experimental Details}\label{app:experimental_details}
In this section we provide the experimental details for all experiments presented in the main paper. Unless otherwise stated, the following parameters are chosen: $\alpha=1$, $\delta=3$, weight decay of $1\times 10^4$ and optimised using SGD with momentum of 0.9. For cross validation experiments, 3 or 5 folds of 10\% of the training data are generated and used for a single run each. Test set experiments use the entire training set and give evaluations on the test sets provided. If no test set is provided then a constant validation set of 10\% of the available data is used. Table~\ref{tab:exp_dets} provides general training details that were present in all experiments.


All experiments were run on a single GTX1080ti or V100, with the exceptions of ImageNet experiments (4 $\times$ GTX1080ti) and DenseNet/PyramidNet experiments (2 $\times$ V100). ResNet18 and LSTM experiments ran within 2 hours in all instances, PointNet experiments ran within 10 hours, WideResNet/DenseNet experiments ran within 2.5 days and auto-augment experiments ran within 10 days.
For all image experiments we use standard augmentations to normalise the image to $[0,1]$ and perform random crops and random horizontal flips. For the google commands experiment we used the transforms and augmentations implemented here \url{https://github.com/tugstugi/pytorch-speech-commands} for their solution to the tensorflow speech recognition challenge.

\begin{table*}
\centering
\caption{General experimental details present in all experiments. Batch Size (BS), Learning Rate (LR). Schedule reports the epochs at which the learning rate was multiplied by 0.1. $^\dagger$ Adam optimiser used. }
\label{tab:exp_dets}
\begin{tabularx}{\textwidth}{XXllll}
\toprule
Experiment & Model & Epochs & Schedule & BS & LR \\
\midrule
\multirow{4}{*}{CIFAR-10 / 100} & PreAct-ResNet18 & 200 & 100, 150 & 128 & 0.1\\
& WideResNet-28-10 & 200 & 100, 150 & 128 & 0.1 \\
& DenseNet-BC-190 & 300 & 100, 150, 225 & 32 & 0.1 \\
& PyramidNet-272-200 & 1800 & Cosine-Annealed & 64 & 0 - 0.05\\
\midrule
\multirow{3}{*}{FashionMNIST} & PreAct-ResNet18 & 200 & 100, 150 & 128 & 0.1\\
& WideResNet-28-10 & 300 & 100, 150, 225 & 32 & 0.1\\
& DenseNet-BC-190 & 300 & 100, 150, 225 & 32 & 0.1\\
\midrule
Google Commands & PreAct-ResNet18 & 90 & 30, 60, 80 & 128 & 0.1\\
\midrule
ImageNet & ResNet101 & 90 & 30, 60, 80 & 256 & 0.4\\
\midrule
TinyImageNet & PreAct-ResNet18 & 200 & 150, 180 & 128 & 0.1\\
\midrule
Bengali.AI & PreAct-ResNet18 & 100 & 50, 75 & 512 & 0.1\\
\midrule
\multirow{3}{*}{Sentiment Analysis$^\dagger$} & CNN & 15 & 10 & 64 & $10^{-3}$ \\
& LSTM & 15 & 10 & 64 & $10^{-3}$ \\
& Bert & 5 & 3 & 32 & $10^{-5}$ \\
\midrule
Combining MSDAs & PreAct-ResNet18 & 200 & 100, 150 & 128 & 0.1 \\
\midrule
ModelNet10$^\dagger$ & PointNet & 50 & 10, 20, 30, 40 & 16 & $10^{-3}$ \\
\midrule
Ablations & PreAct-ResNet18 & 200 & 100, 150 & 128 & 0.1\\
\bottomrule
\end{tabularx}
\end{table*}


\bibliographystyle{IEEEtranN}

\bibliography{main}
\end{document}